\documentclass[10pt,twocolumn,letterpaper]{article}

\usepackage{cvpr}
\usepackage{times}
\usepackage{epsfig}
\usepackage{graphicx}
\usepackage{amsmath}
\usepackage{amssymb}
\usepackage{bbm}
\usepackage{bm}
\usepackage{authblk}
\usepackage{bbding}
\usepackage{color}

\newcommand{\hj}[1]{{\color{red}{#1}}}

\newcommand{\tabincell}[2]{\begin{tabular}{@{}#1@{}}#2\end{tabular}}

\usepackage[pagebackref=true,breaklinks=true,letterpaper=true,colorlinks,bookmarks=false]{hyperref}

\cvprfinalcopy 

\ifcvprfinal\fi
\begin{document}

\title{G2L-Net: Global to Local Network for Real-time 6D Pose Estimation with Embedding Vector Features}

\author{Wei Chen $^{1,2}$  \hspace{0.7cm} Xi Jia$^{1}$ \hspace{0.7cm} Hyung Jin Chang$^{1}$\hspace{0.7cm} Jinming Duan$^{1}$\hspace{0.7cm}  Ales Leonardis$^{1}$\\
{$^1$ School of Computer Science, University of Birmingham  \\ $^2$ School of Computer Science, National University of Defense Technology}\\
{\tt\small \{\Envelope wxc795,X.Jia.1,h.j.chang,j.duan,a.leonardis\}@cs.bham.ac.uk}
}



\maketitle
\thispagestyle{empty}
\begin{abstract}
In this paper, we propose a novel real-time 6D object pose estimation framework, named G2L-Net. Our network operates on point clouds from RGB-D detection in a divide-and-conquer fashion. Specifically, our network consists of three steps. First, we extract the coarse object point cloud from the RGB-D image by 2D detection. Second, we feed the coarse object point cloud to a translation localization network to perform 3D segmentation and object translation prediction. Third, via the predicted segmentation and translation, we transfer the fine object point cloud into a local canonical coordinate, in which we train a rotation localization network to estimate initial object rotation. In the third step, we define point-wise embedding vector features to capture viewpoint-aware information. To calculate more accurate rotation, we adopt a rotation residual estimator to estimate the residual between initial rotation and ground truth, which can boost initial pose estimation performance. Our proposed G2L-Net is real-time despite the fact multiple steps are stacked via the proposed coarse-to-fine framework. Extensive experiments on two benchmark datasets show that G2L-Net achieves state-of-the-art performance in terms of both accuracy and speed. \footnote{Our code is available at \url{https://github.com/DC1991/G2L_Net}.}



\end{abstract}

\section{Introduction}
\label{sec:intro}

Real-time performance is important in many computer vision tasks, such as, object detection \cite{redmon2018yolov3,liu2016ssd}, semantic segmentation \cite{ren2015faster, he2017mask}, object tracking \cite{chen2016appearance, henriques2014high}, and pose estimation \cite{peng2018pvnet,Tekin_2018_CVPR,Kehlssd}. In this paper, we are interested in real-time 6D object pose estimation, which has significant impacts on augmented reality \cite{marchand2016pose, marder2016project}, smart medical and robotic manipulation \cite{zhu2014single,tremblay2018deep}.

 \begin{figure}[t!]
\centering
\includegraphics[width=0.48\textwidth]{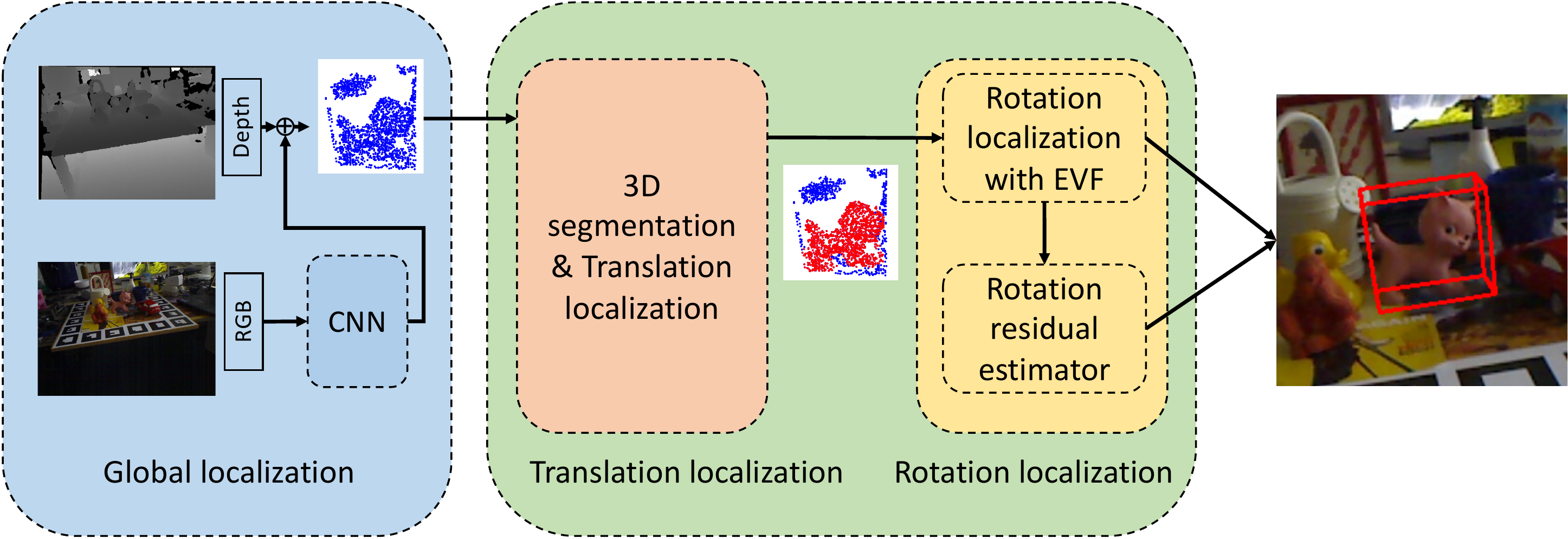}
\caption{\textbf{Three steps of G2L-Net.} We propose a novel real-time point cloud based network for 6D object pose estimation called G2L-Net. Our G2L-Net contains global localization, object translation localization and rotation localization. For the rotation localization, we propose point-wise embedding vector features (EVF) and rotation residual estimator to access accurate rotation.}
\label{fig:first_over}
\end{figure} 
Deep learning methods have shown the state-of-the-art performance in the pose estimation tasks, but many of them \cite{Xiang2017,Rad2017bb8,Oberweger2018,Li_2018_ECCV} cannot run in real-time. While there exist some real-time deep learning methods \cite{peng2018pvnet,Rad2018,Tekin_2018_CVPR,zakharov2019dpod} ($>$ 20fps), they use only RGB information from an image. One major limitation of using RGB only is that features learned from such information are sensitive to occlusion and illumination changes, which precludes these methods from being applied to complicated scenes. Deep learning methods based on depth information \cite{Kehl2016, Li_2018_ECCV} are more suitable for realistically complicated scenes, but they are usually computation-intensive.
One common issue for these RGBD-based methods is that exploiting viewpoint information from depth information is not very effective, thus reducing their pose estimation accuracy. To overcome this, these methods tend to use a post-refinement mechanism or a hypotheses generation/verification mechanism to enhance pose estimation accuracy. This, however, reduces the inference speed for pose estimation.


{In this paper,} to overcome the existing problems in depth-based methods, we propose a global to local real-time network (G2L-Net), with two added modules which are point-wise embedding vector features extractor and rotation residual estimator. 
Built on \cite{Qi_2018_CVPR}, our method has three major novelties: i) instead of locating the object point cloud by a frustum, we locate the object point cloud by a 3D sphere, which can limit the 3D search range in a more compact space (see Section \ref{sec:global_loc} for details), ii) instead of directly regressing the global point feature to estimate the pose, we propose the point-wise embedding vector features to effectively capture the viewpoint information, and iii) we estimate the rotation residual between predicted rotation and the ground truth. The rotation residual estimator further boosts the pose estimation accuracy.
We evaluate our method on two widely-used 6D object pose estimation datasets, \ie LINEMOD \cite{Hinterstoisser2013} and YCB-Video \cite{Xiang2017} dataset. Experimental results show that G2L-Net outperforms state-of-the-art depth-based methods in terms of both accuracy and speed on the LINEMOD dataset, and that G2L-Net achieves comparable accuracy while is the fastest method on the YCB-Video dataset. 

In summary, the main contributions of this paper are as follows:
\begin{itemize}
    \item We propose a novel real-time framework to estimate 6D object pose from RGB-D data in a global to local (G2L) way. Due to efficient feature extraction, the framework runs at over 20fps on a GTX 1080 Ti GPU, which is fast enough for many applications.
    
    \item We propose orientation-based point-wise embedding vector features (EVF) which better utilize viewpoint information than the conventional global point features.
    \item We propose a rotation residual estimator to estimate the residual between predicted rotation and ground truth, which further improves the accuracy of rotation prediction.
\end{itemize}

\section{Related work}
\noindent
\textbf{Pose estimation from RGB image:} Traditional methods \cite{Hinterstoisser2013,lepetit2005monocular, lowe1999object} compute 6D object pose by matching RGB features between a 3D model and test image. These methods use handcrafted features that are not robust to background clutter and image variations \cite{yuan2016congested, shin2017comparison,peng2018pvnet}. 
Learning-based methods \cite{Rad2017bb8, peng2018pvnet, Rad2018,Oberweger2018,Hu_2019_CVPR,Tekin_2018_CVPR} alleviate this problem by training their model to predict 2D keypoints and compute the object pose by the PnP algorithm \cite{Gao2003, Morenonoguer2007}. \cite{Xiang2017,LiIm,Li_2018_ECCV} decouple the pose estimation into two sub-tasks: translation estimation and rotation estimation. More concretely, they regarded the translation and rotation estimation as a classification problem and trained neural networks to classify the image feature into a discretized pose space. However, the RGB image features may be affected by illumination changes which result in pose estimation from RGB image more sensitive to illumination changes. 

\noindent
\textbf{Pose estimation from RGB image with depth information:}
When depth information is available, previous approaches \cite{Brachmann2014, Tejani2014, wohlhart2015learning, hinterstoisser2016going} learned features from the input RGB-D data and adopted correspondence grouping and hypothesis verification. 
However, some papers \cite{yuan2016congested, shin2017comparison} found that the methods are sensitive to image variations and background clutter. Besides, correspondence grouping and hypothesis verification further increase the inference time presenting real-time applications.
Some methods \cite{Xiang2017, Kehlssd} employed the depth information in a post-refinement procedure by highly customized Iterative Closest Point (ICP) \cite{chen1992object,besl1992method} into deep learning frameworks, which would significantly increase the running time of the algorithms.
Recently, several deep learning methods \cite{Kehl2016,Li_2018_ECCV} utilized the depth input as an extra channel along with the RGB channels. However, combining depth and RGB information in this way cannot make full use of geometric information in data and makes it difficult to integrate information across viewpoints \cite{maturana2015voxnet}. Instead, in this paper, we transfer depth maps to 3D point clouds and directly process the 3D point clouds by PointNets \cite{Qi_2017_CVPR, Qi_2018_CVPR} which extract 3D geometric features more efficiently than CNN-based architectures.

\noindent
\textbf{Pose estimation from point cloud:}
PointNets \cite{Qi_2017_CVPR, Qi_2018_CVPR}, Qi \textit{et al.} have shown that employing depth information in 3D space via the point cloud representation could achieve better performance than that in 2.5D space. Based on that, some PointNet-based methods \cite{Qi_2018_CVPR, wang2019densefusion, zhou2018voxelnet,yang2018pixor, chen2020ponitposenet} presented to directly estimate 6D object pose. They adopted a PoinNet-like \cite{Qi_2017_CVPR} architecture to access pose from point cloud. In this work, we also make use of PointNet-like architecture but in a different way. Different from 2D methods \cite{Xiang2017,Li_2018_ECCV}, we decouple 6D object pose estimation into three sub-tasks: global localization, translation localization, and rotation localization. For the first two sub-tasks, we use similar methods in \cite{Qi_2018_CVPR} but with some improvements described in Section \ref{sec:proposed}.
For the third sub-task, we propose point-wise embedding vector features that exploit the viewpoint information more effectively and we also propose a rotation residual estimator that further improves the pose estimation accuracy. We show that with these improvements, the proposed G2L-net achieves higher accuracy than state-of-the-art methods and runs at real-time speed.

\begin{figure*}[ht!]
\centering
\includegraphics[width=0.99\textwidth]{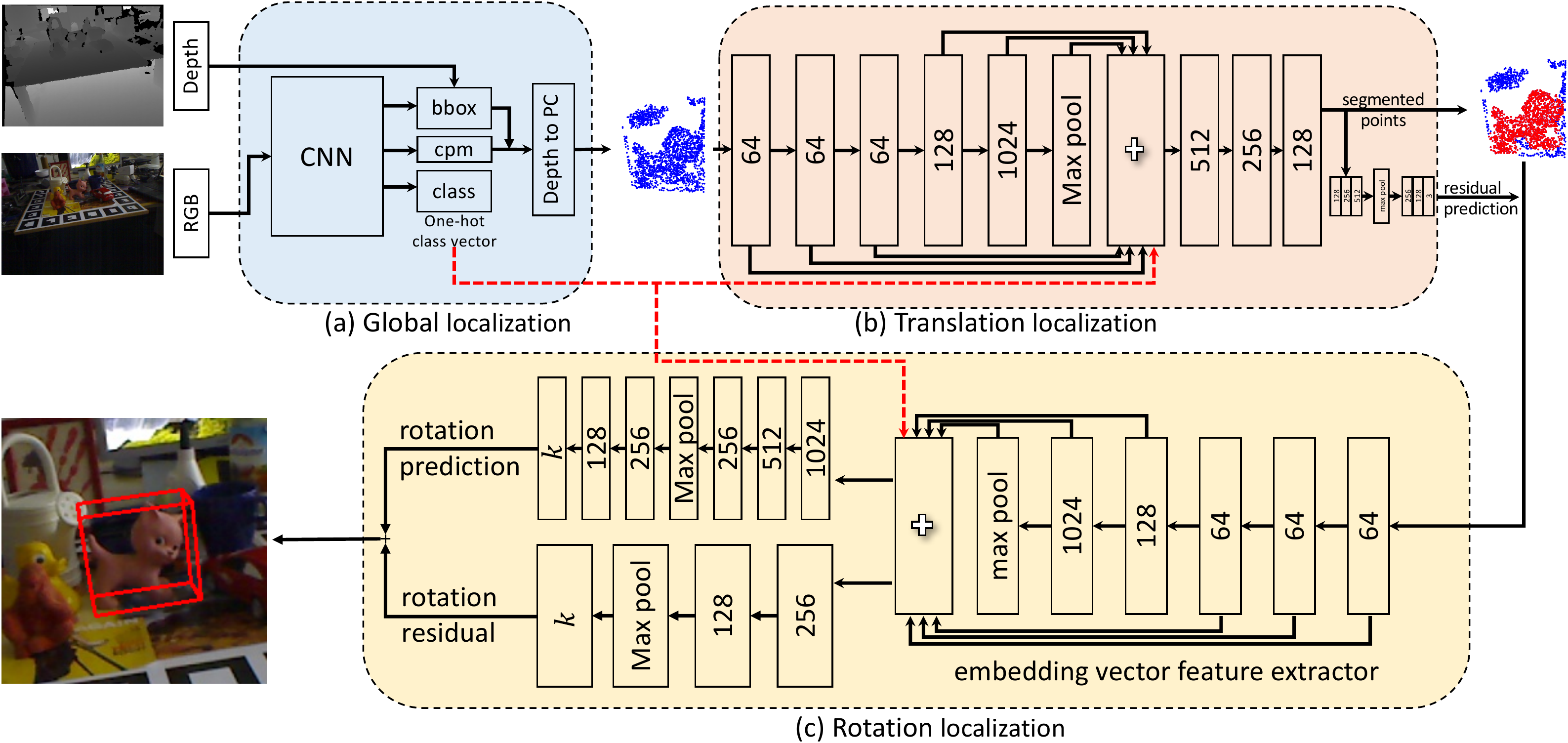}
\vspace{10pt}
\caption{\textbf{Inference pipeline of the proposed G2L-Net.} (a) For RGB image, we use a 2D detector to detect the bounding box (bbox) of the target object and the object label which is used as a one-hot feature for the following networks. Also, we additionally choose the maximum probability location in class probability map (cpm) as the sphere center (we transfer this 2D location to 3D with known camera parameters and corresponding depth value) which is used to further reduce the 3D search space. (b) Given the point clouds in the object sphere, we use a translation localization network to perform 3D segmentation and translation residual prediction. Then we use the 3D segmentation mask and the predicted translation to transfer the object point cloud into a local canonical coordinate. (c) In the rotation localization network, we first use the point-wise embedding vector feature extractor to extract embedding vector features. Then we feed this feature into two-point clouds decoders: the top decoder directly outputs the rotation of the input point cloud and the bottom one outputs the residual of the output of the top one between the ground truth. $k$ is the dimension of the output vector.$``+''$ denotes feature concatenation.}
\label{fig:whoarch}
\end{figure*} 

\section{Proposed method}
\label{sec:proposed}

In Figure \ref{fig:whoarch}, we show the inference pipeline of our proposing G2L-Net which estimates the 6D object pose in three steps: global localization, translation localization, and rotation localization. In the global localization step, we use a 3D sphere to fast locate the object position in 3D space. In the translation localization step, we train a PointNet to perform 3D segmentation and estimate object translation. In the third step, we estimate rotation with the proposed point-wise embedding vector features and rotation residual estimator. Please note, this rotation residual estimator is different from the post-refinement component in previous methods \cite{Xiang2017,Kehlssd}, it outputs rotation residual with initial rotation synchronously.
In the following subsections, we describe each step in detail.

\subsection{Global localization}
\label{sec:global_loc}
\begin{figure}
\begin{center}
\begin{tabular}{cc}
{{\includegraphics[width=0.2\textwidth]{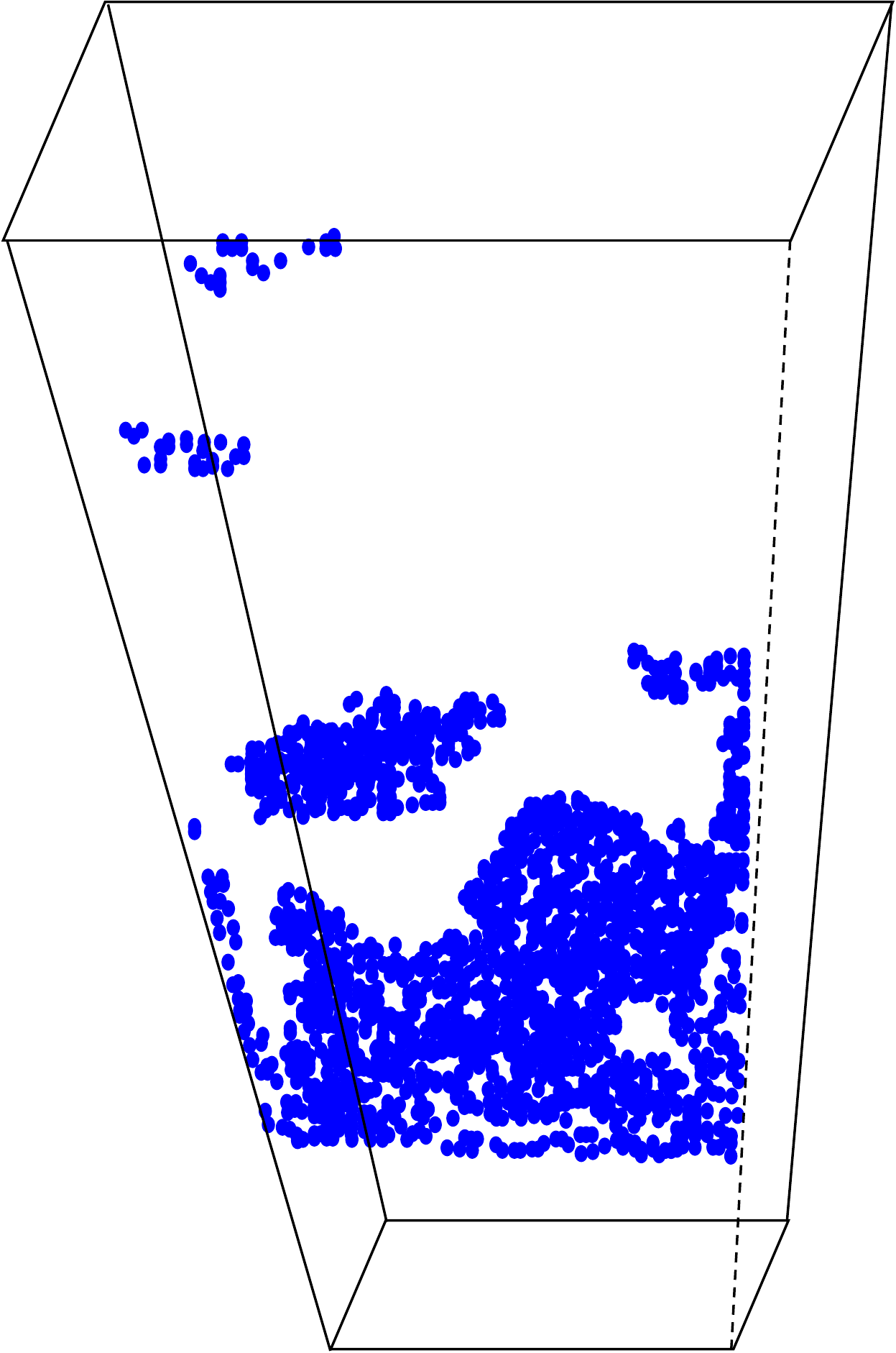}}}&
{{\includegraphics[width=0.2\textwidth]{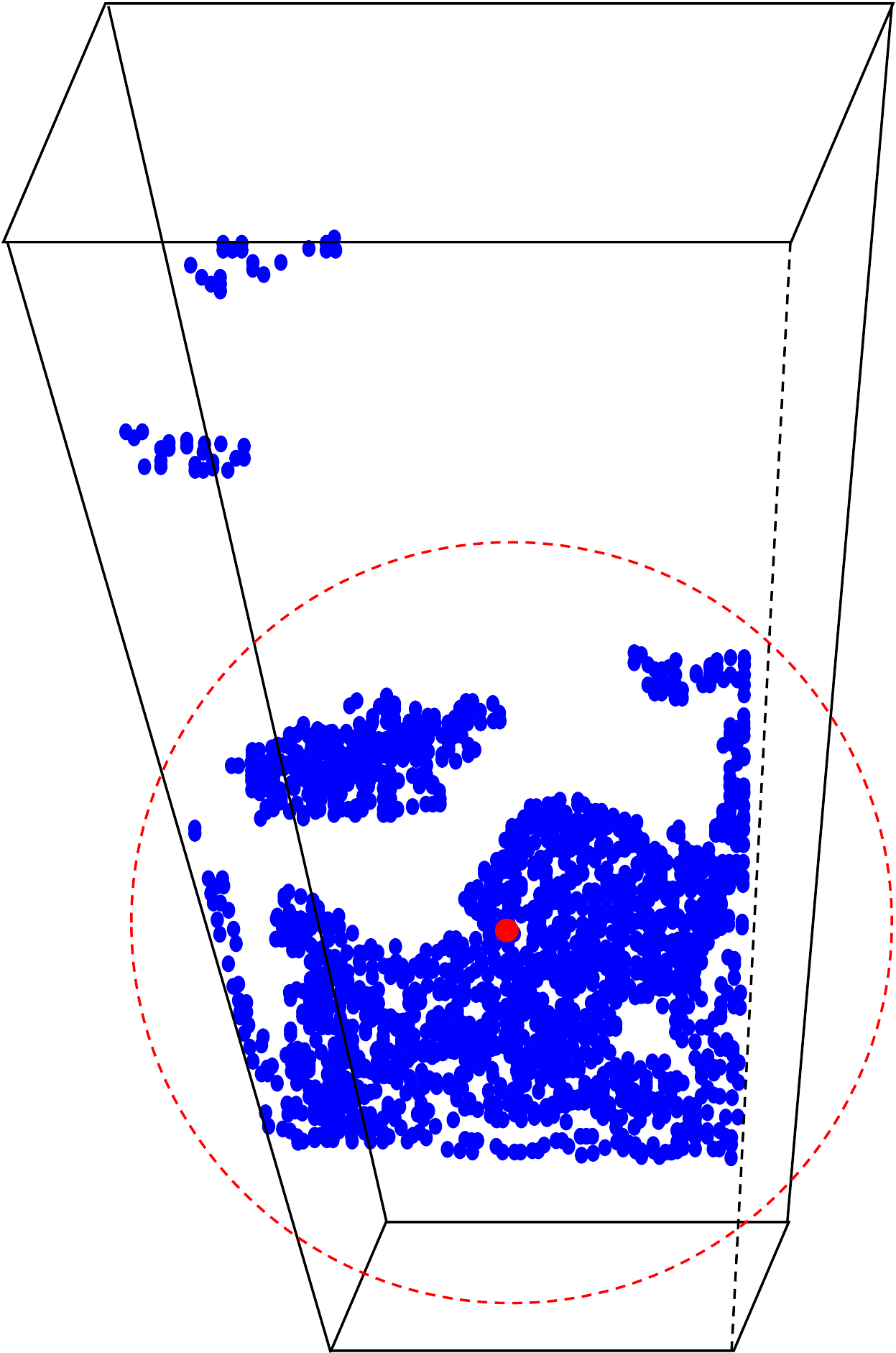}}}\\
(a)&(b)\\
\end{tabular}
\end{center}
\vspace{5pt}
\caption{\textbf{Global 3D sphere.} In the global localization step, we locate the object point clouds by bounding box as well as a 3D sphere. (a) Locate the object point cloud only by bounding box. In this case, it can only locate the object in two-dimensional space, some points can still very far away from the object in the third axis. (b) Locate the object point cloud by both the bounding box and 3D sphere. All points lay in a more compact space.}
\label{fig:sphere}
\end{figure}

To fast locate the global position of the target object in the whole scene, we train a 2D CNN detector, YOLO-V3 \cite{redmon2018yolov3}, to detect the object bounding box in RGB image, and output object label which is used as one-hot class vector for better point cloud instance segmentation, translation and rotation estimation.
In \cite{Qi_2018_CVPR}, they use the 2D bounding box to generate frustum proposals which only reduce the 3D search space of two axes ($x$,$y$). Differently, rather than only using a 2D bounding box, we propose to employ a 3D sphere to further reduce the 3D search space in the third axis ($z$) (see Figure \ref{fig:sphere} for details). The center of the 3D sphere is transferred from the 2D location which has the maximum value in the class probability map with known camera parameters and corresponding depth value. The radius of this 3D sphere is the diameter of the detected object. We only choose points in this compact 3D sphere, which makes the learning task easier for the following steps.

\subsection{Translation localization}
Although the extracted point cloud is tight, there are still two issues remained: 1) the point cloud in this 3D space contains both object points and non-object points, and 2) the object points cannot be transferred to a local canonical coordinate due to unknown translation.
To cope with the issues, similar to \cite{Qi_2018_CVPR}, we train a two PointNets \cite{Qi_2017_CVPR} to perform 3D segmentation and output the residual distance $||T-\Bar{T}||_2$ between the mean value $\Bar{T}$ of the segmented points and object translation $T$. This residual can be used to calculate the translation of the object. 

\subsection{Rotation localization with embedding vector feature}
\begin{figure}[t!]
\centering
\begin{tabular}{c}
{{\includegraphics[width=0.3\textwidth]{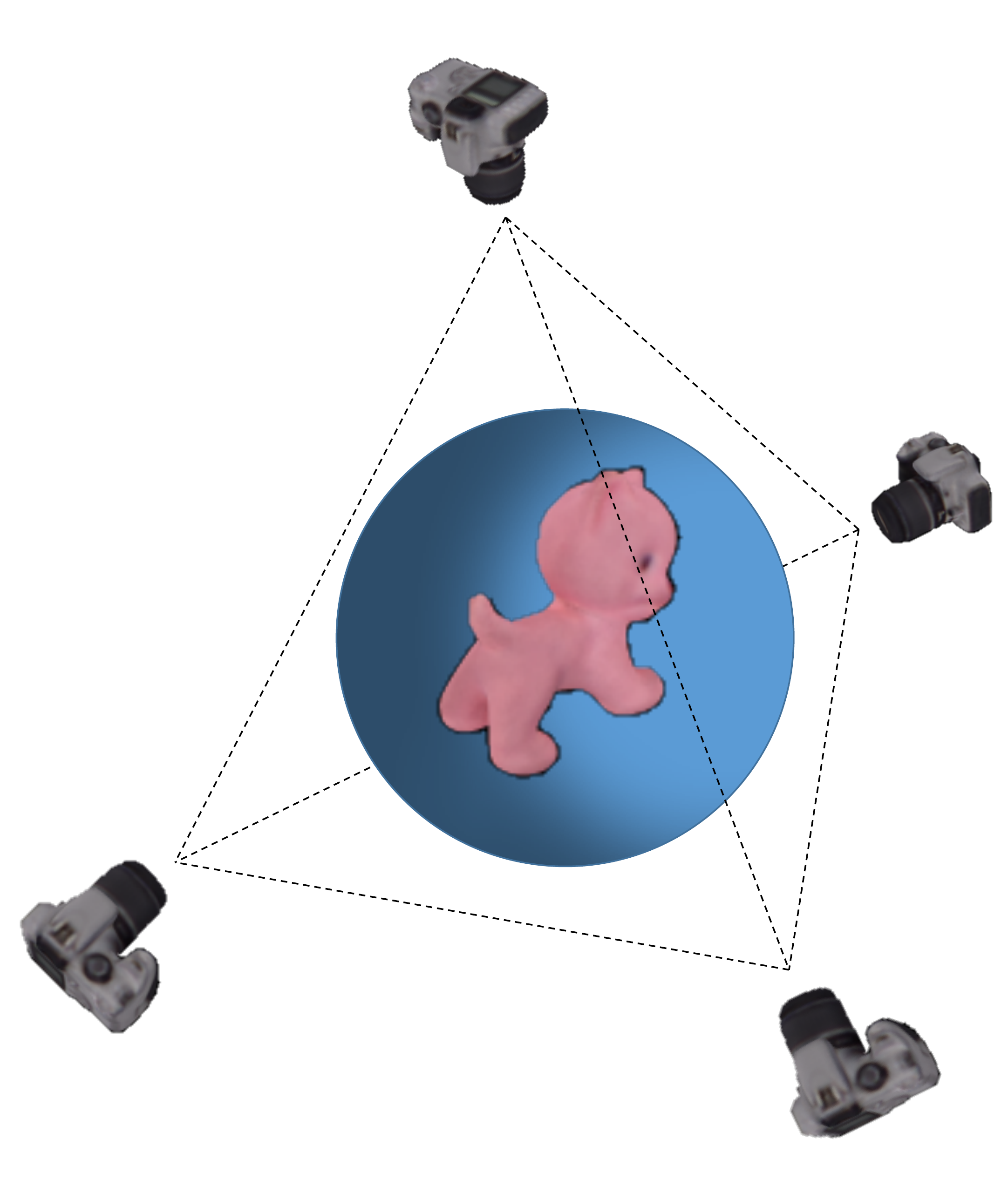}}}
\end{tabular}
\caption{\textbf{Different viewpoints.} For a 3D object, we need at least four viewpoints to cover all the points of the 3D object.}
\label{fig:views}
\vspace{-10pt}
\end{figure}

From the first two steps, we transfer the point cloud of the object to a local canonical space where the viewpoint information is more evident.
Theoretically, we need at least four different viewpoints to cover all points of an object (see Figure \ref{fig:views}) in 3D space \cite{draim1987common}. For the pose estimation task, we usually have hundreds of different viewpoints for one object during training. 
Then our goal is to use the viewpoint information adequately. In \cite{Qi_2018_CVPR}, they use PointNets \cite{Qi_2017_CVPR,qi2017pointnetplusplus} to extract the global feature from the whole point cloud. However, in our experiments, we found global point features extracted from point clouds under similar viewpoints are highly correlated, which limit the generalization performance (see Figure \ref{fig:gen_per} in the experiment section). 

\begin{figure}
\begin{center}
\begin{tabular}{c}
{{\includegraphics[width=0.45\textwidth]{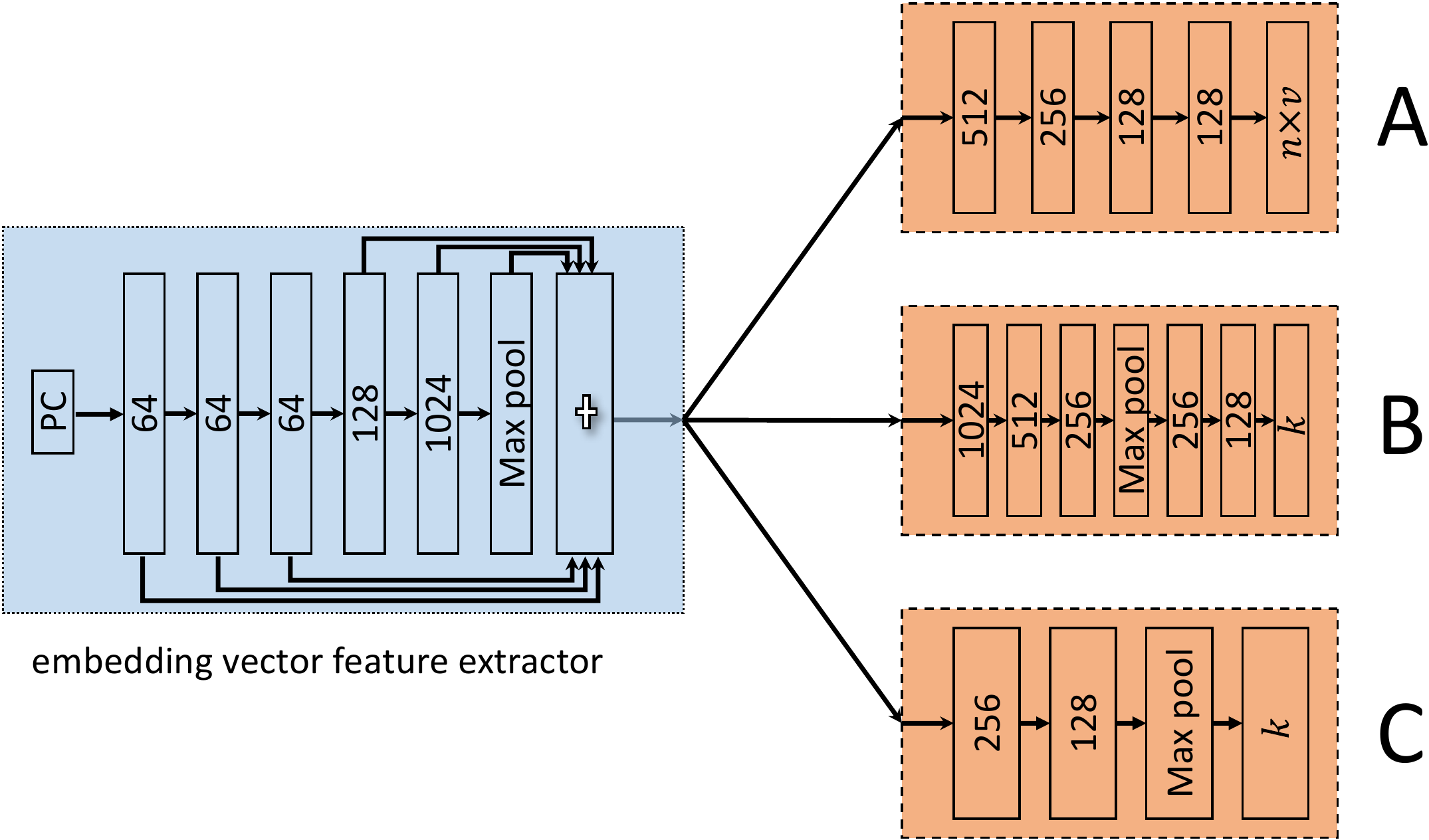}}}\\
\end{tabular}
\end{center}
\caption{\textbf{Architecture of the rotation localization network.} In the training stage, there are three blocks in rotation localization network. We train block A to predict the unit vectors pointing to the keypoints, the loss function of this block is the mean square error between the predicted and ground truth directional vectors.
By training this block, the network can learn how to extract point-wise embedding vector features from the input point cloud. Note that, block A is not deployed in the inference stage.
Then we use block B to integrate the point-wise embedding vector features to predict object rotation. The loss function of this block is the mean square error between the predicted rotation and ground truth. For rotation residual estimator block C, we use the Euclidean distance between the predicted 3D keypoints position (output of block B) and ground truth as ground truth. $k$ is the dimension of the output rotation vector and $v$ is the dimension of the output directional vector. $``+''$ denotes feature concatenation.
}
\label{fig:rotation_net}
\end{figure}

\begin{figure}[t!]
\centering
\begin{tabular}{c}
{{\includegraphics[width=0.3\textwidth]{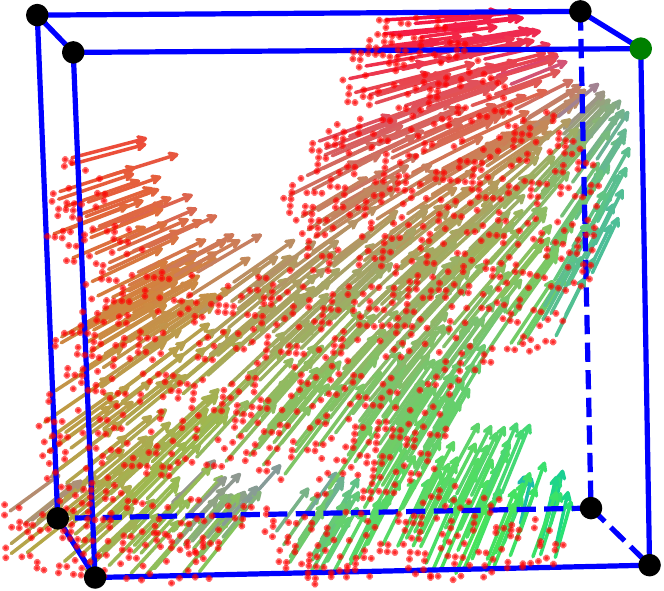}}}
\end{tabular}
\caption{\textbf{Point-wise vectors.} Here we show point-wise vectors pointing to one keypoint which is shown in green color, and other keypoints are shown in black color. We train our network to predict such directional vectors}
\label{fig:vectors}
\end{figure}

\begin{figure}
\begin{center}
\begin{tabular}{cc}
{{\includegraphics[width=0.21\textwidth]{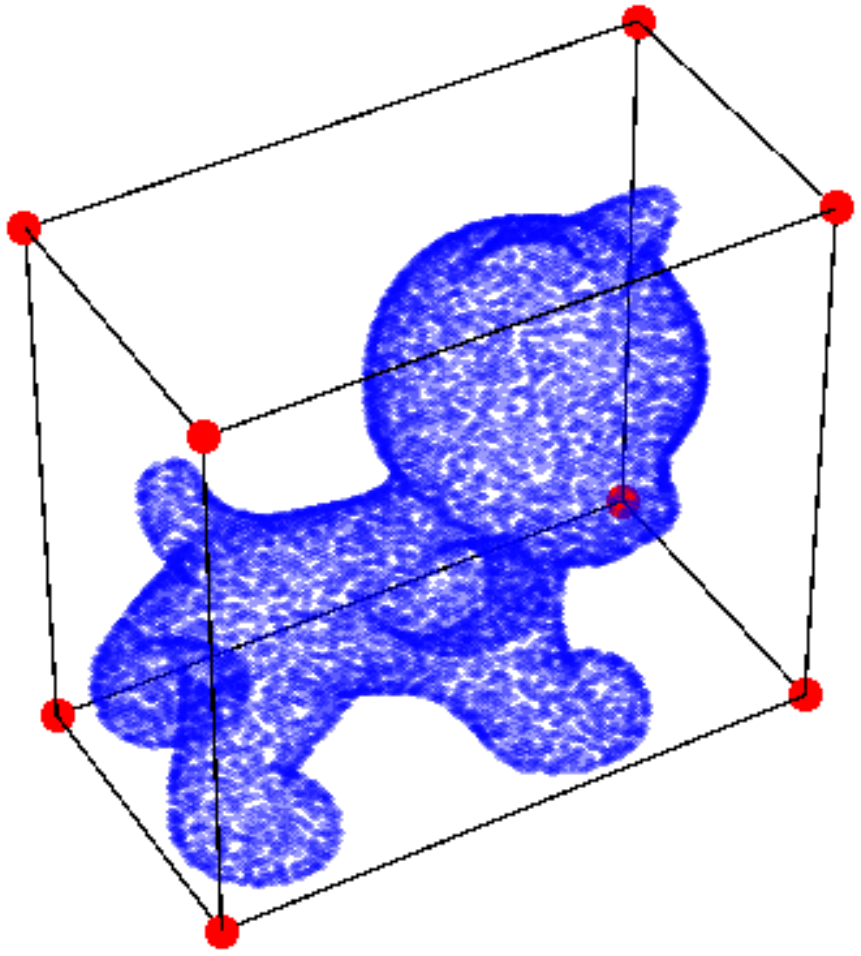}}}&
{{\includegraphics[width=0.16\textwidth]{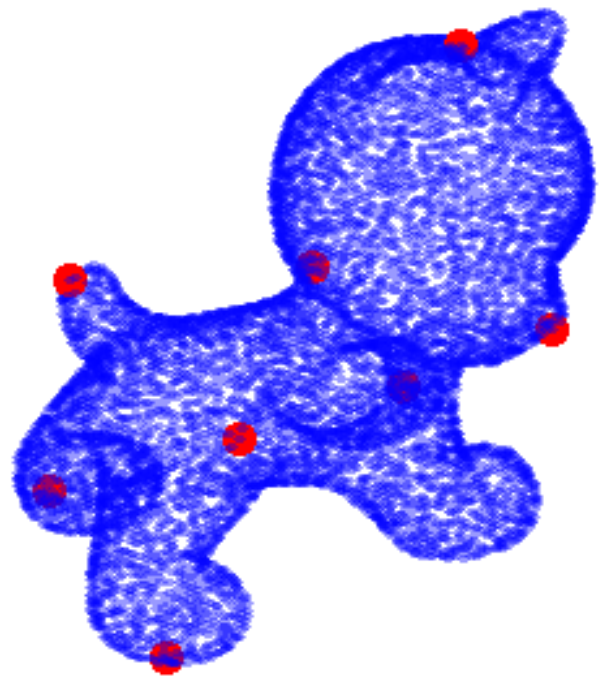}}}\\
(a)&(b)\\
\end{tabular}
\end{center}
\vspace{5pt}
\caption{\textbf{Visualization of different keypoint selection schemes}. The left image is a 3D object point cloud and its 3D bounding box; the right image is the keypoint selected by FPS algorithm. The keypoints are shown in red color.}
\label{fig:kps_slec}
\end{figure}

To overcome the limitation of global point features, we propose point-wise embedding vector features.
Specifically, we design the rotation localization network architecture as shown in Figure \ref{fig:rotation_net} to predict point-wise unit vectors pointing to keypoints (illustrated in Figure \ref{fig:vectors}). The keypoints are some pre-defined 3D points based on each 3D object model. Two aspects need to be decided for the keypoints: number and location. A simple way is to use the 8 corners of the 3D bounding box of object model as keypoints which we show in Figure \ref{fig:kps_slec} (a). This definition is widely used by many CNN based methods in 2D cases \cite{Rad2017bb8, Rad2018, Oberweger2018,Tekin_2018_CVPR}. 
Another way is, as proposed in \cite{peng2018pvnet}, to use the farthest point sampling (FPS) algorithm to sample the keypoints in each object model. Figure \ref{fig:kps_slec} shows examples of different keypoint selection schemes. In Section \ref{sec:abla}, we show how the number and location of the keypoints influence the pose estimation results. 

Similar to \cite{chen2020ponitposenet}, our proposed rotation localization network takes object point cloud in the local canonical space and outputs point-wise unit vectors pointing keypoints. The loss function is defined as follows:
\begin{equation}
\label{eq:lossv}
\ell(\bm{\theta})=\min_{\bm{\theta}}\frac{1}{K \left |\mathcal{X} \right |}\sum_{k=1}^{K}
\sum_{i} \left \|  \widetilde{v}_k(\mathcal{X}_i; {\bm{\theta}})-v_k(\mathcal{X}_i) \right \|_2^2,
\end{equation}
where $K$ is the number of keypoints. $\bm{\theta}$ is the network parameters. $\widetilde{v}_k(\mathcal{X}_i; \bm{\theta})$ and $v_k(\mathcal{X}_i)$ are the predicted vectors and the ground truth, respectively. $\mathcal{X}\in \mathbbm{R}^{n\times3}$ represents the object points in the local coordinate space. $|\mathcal{X}|$ is the number of object points.


Different from other state-of-the-art methods \cite{peng2018pvnet,Xiang2017,chen2020ponitposenet}, we adopt a multilayer perceptron (MLP) that takes point-wise embedding vector features as input and outputs the rotation of object as shown in Figure \ref{fig:rotation_net}. Please note, during inference, we use the rotation matrix to represent the rotation which is computed from the keypoint positions using the Kabsch algorithm. Over the training process, as per the definition of point-wise vectors, we used only the keypoint positions to represent rotation.
In experiments, we have found that our proposed method can make faster and more accurate predictions than the methods \cite{peng2018pvnet,Xiang2017,chen2020ponitposenet}.


\textbf{Rotation residual estimator: }
To better utilize the viewpoint information in the point-wise embedding vector features, we add an extra network branch (block C in Figure \ref{fig:rotation_net}) to estimate the residual between estimated rotation (block B in Figure \ref{fig:rotation_net}) and ground truth. 
However, we do not have ground truth for this residual estimator. To address this problem we train this estimator in online fashion. Assuming that the ground truth for block B of rotation localization network is $\mathcal{P}$ and the output of block B is $\widetilde{\mathcal{P}}$, then the target of our rotation residual estimator is $||\mathcal{P} - \widetilde{\mathcal{P}}||_2$. 
As the rotation network converging, it becomes harder to learn the residual. If the rotation localization network can fully exploit the embedding vector feature, the role of rotation residual estimator can be ignored. However, when the rotation network cannot fully exploit the embedding vector feature, the rotation residual estimator will have a big impact on the final results, we show this property of rotation residual estimator in Figure \ref{fig:gen_per} (b). Please note, our proposed rotation residual estimator is different from the post-refinement module in the previous state-of-the-art methods \cite{Xiang2017,wang2019densefusion,LiIm}. Our proposed rotation residual estimator outputs rotation residual with estimated rotation synchronously, which saves the running time.


\section{Experiments}
There are two parts in this experiments section. Firstly, we do ablation studies on keypoints selection schemes and empirically validate the three innovations introduced in our new frame: 3D sphere (``SP”), point-wise embedding vector features (``EVF”) and rotation residual estimator (``RRE"). Then we test our proposed G2L-Net on two benchmark datasets, \ie  LINEMOD and YCB-Video datasets. Our method achieves state-of-the-art performance in real-time on both datasets. 

\subsection{Implementation details}
We implement our framework using Pytorch. We have conducted all the experiments on an Intel i7-4930K 3.4GHz CPU with one GTX 1080 Ti GPU.
First, we fine-tune the YOLO-V3 \cite{redmon2018yolov3} architecture which is pre-trained on the ImageNet \cite{deng2009imagenet} to locate the 2D region of interest and access the class probability map.
Then we jointly train our proposed translation localization and rotation localization networks using PointNet \cite{Qi_2017_CVPR} as our backbone network. The architectures of these networks are shown in Figure \ref{fig:whoarch}. Note that, other point-cloud network architectures \cite{qi2017pointnetplusplus, zhou2018voxelnet} can also be adopted as our backbone network.
For point cloud segmentation we use cross-entropy as the loss function. For translation residual prediction, we employ the mean square error and the unit in our experiment is $mm$. We train our rotation localization network as described in Figure \ref{fig:rotation_net}. 
We use Adam \cite{kingma2014adam} to optimize the proposed G2L-Net. We set the initial learning rate as 0.001 and halve it every 50 epochs. The maximum epoch is 200.

\subsection{Datasets}
\noindent
\textbf{LINEMOD} \cite{Hinterstoisser2013} is a widely used dataset for 6D object pose estimation. There are 13 objects in this dataset. For each object, there are about 1100-1300 annotated images and each has only one annotated object. This dataset exhibits many challenges for pose estimation: texture-less objects, cluttered scenes, and lighting condition variations.\\
\textbf{YCB-Video} \cite{Xiang2017} contains 92 real video sequences for 21 YCB object instances \cite{calli2015benchmarking}. This dataset is challenging due to the image noise and occlusions.

\begin{figure}[t!]
\centering
\begin{tabular}{ccc}
{{\includegraphics[width=0.14\textwidth]{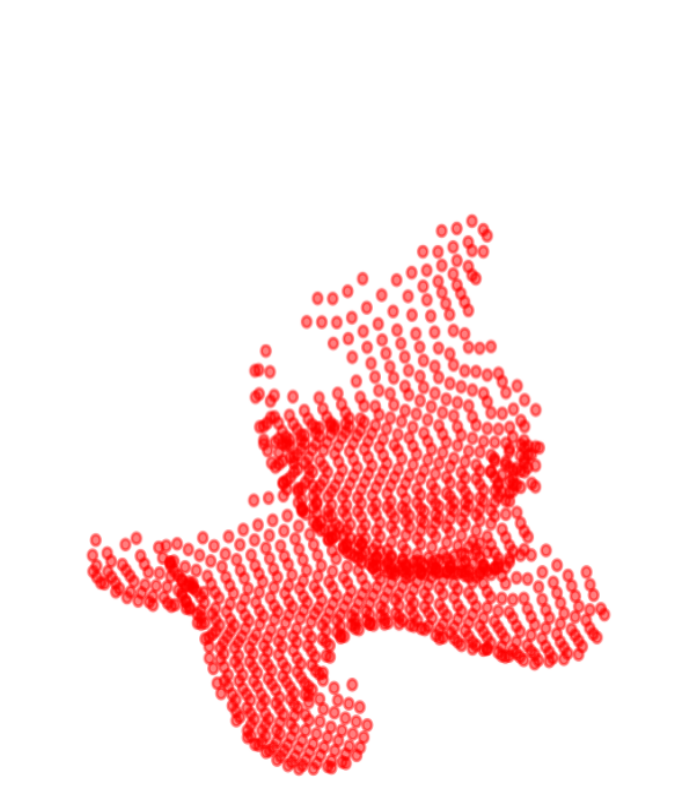}}}&
{{\includegraphics[width=0.14\textwidth]{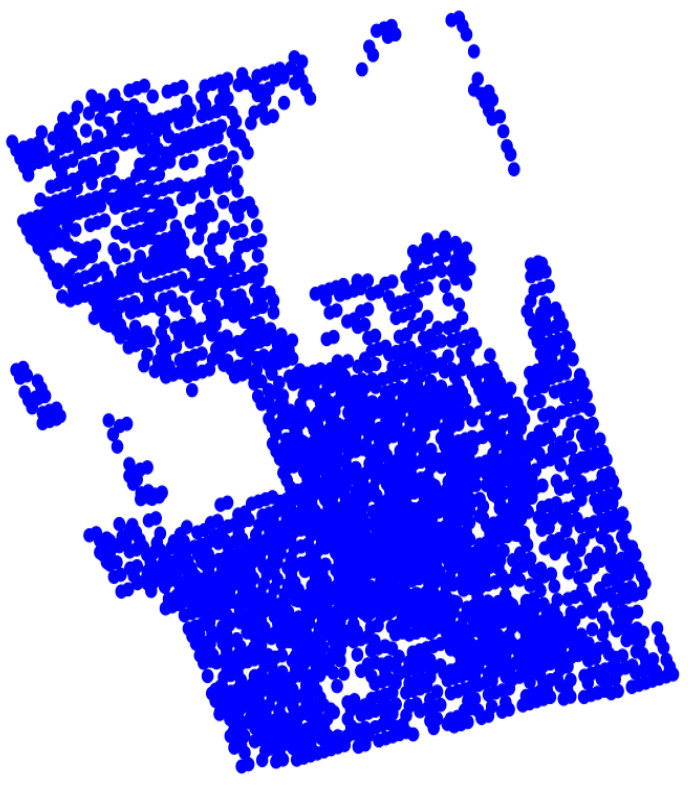}}}&
{{\includegraphics[width=0.14\textwidth]{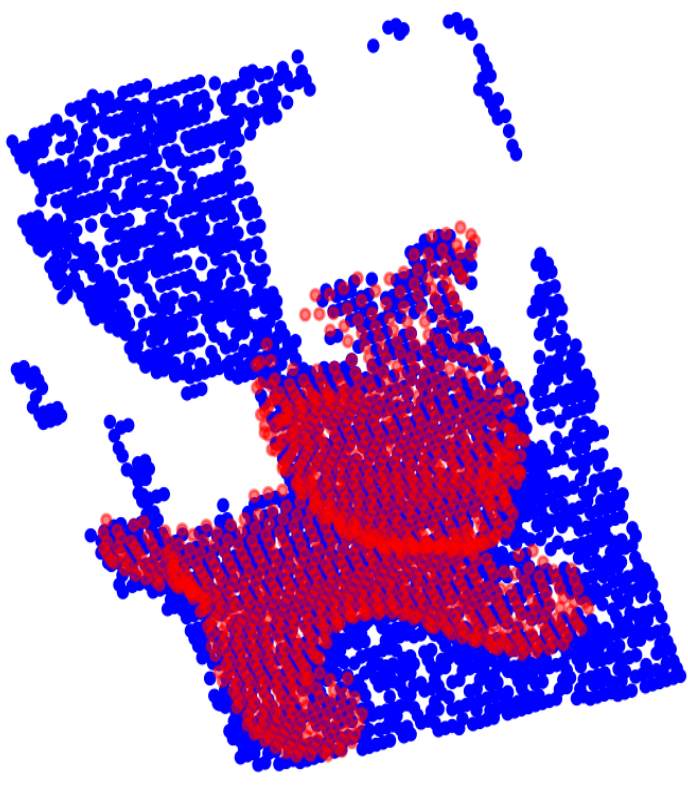}}}\\
(a)&(b)&(c)\\
\end{tabular}
\caption{\textbf{Point cloud labeling}. (a) The object model of $cat$ in LINEMOD dataset;
(b) the point cloud from the depth images in object region; (c) the transformed object model is overlapped on the point cloud. We label each point according to the shortest distance between the point and the corresponding transformed object model.}
\label{fig:label_pro}
\end{figure}

However, both LINEMOD and YCB-Video datasets do not contain the label for each point of the point cloud. To train G2L-Net in a supervised fashion, we adopt an automatic way to label each point of the point cloud of \cite{chen2020ponitposenet}. As described in \cite{chen2020ponitposenet}, we label each point in two steps§
First, for the 3D model of an object, we transform it into the camera coordinate using the corresponding ground truth. We adopt the implementation provided by \cite{hodavn2016evaluation} for this process. Second, for each point on the point cloud in the target region, we compute its nearest distance to the transformed object model. If the distance is less than a value $\epsilon=8mm$, we label the point as 1 (belonging to the object), otherwise 0. Figure \ref{fig:label_pro} shows the labeling procedure.

\subsection{Evaluation metrics}
We employ the ADD metric \cite{Hinterstoisser2013} to evaluate our G2L-Net on LINEMOD dataset:
\begin{equation}
        \frac{1}{\left | \mathcal{M} \right |} \sum _{\textbf{x}\in \mathcal{M}}  \| (\textbf{R}\cdot \textbf{x}+\textbf{T}) - (\widetilde{\textbf{R}} \cdot \textbf{x}+\widetilde{\textbf{T}})  \|,
\end{equation}
where $\left | \mathcal{M} \right |$ is the number of points in the object model. $\textbf{x}$ represents the point in object 3D model. $\textbf{R}$ and $\textbf{T}$ are the ground truth \hj{pose}, and  $\widetilde{\textbf{R}}$ and $\widetilde{\textbf{T}}$ are the estimated pose. In this metric, the mean distance between the two transformed point sets is computed. When the average distance is less than 10\% of the 3D object model diameter, we consider that estimated 6D pose as correct. For symmetric objects, we employ ADD-S metric \cite{Hinterstoisser2013}, where the average distance is calculated using the shortest point distance:
\begin{equation}
     \frac{1}{\left | \mathcal{M} \right |} \sum _{\textbf{x}_1\in \mathcal{M}} \min_{{\textbf{x}_2\in \mathcal{M}}}  \| (\textbf{R}\cdot \textbf{x}_1+\textbf{T}) - (\widetilde{\textbf{R}} \cdot \textbf{x}_2+\widetilde{\textbf{T}})  \|.
\end{equation}

When evaluating on YCB-Video dataset, same as \cite{Xiang2017,peng2018pvnet,Li_2018_ECCV}, we use the ADD-S AUC metric proposed in \cite{Xiang2017}, which is the area under the accuracy-threshold curve. The maximum threshold is set to 10cm \cite{Xiang2017}.

\subsection{Ablation studies}
\label{sec:abla}
Compared to the baseline method \cite{Qi_2018_CVPR}, our proposed method has three novelties. First, we fast locate the object point clouds by a 3D sphere which is different from the frustum method in \cite{Qi_2018_CVPR}. Second, we use the proposed point-wise embedding vector features to estimate rotation of the point cloud which can better utilize the viewpoint information. Third, we propose a rotation residual estimator to estimate the rotation residual between ground truth and predicted rotation.
From Table \ref{tab:abla_evf}, we can see that the proposed three improvements can boost performance.

We also compare the different keypoints selection schemes in Table \ref{tab:abla_kps}, however, it shows that different keypoints selection schemes make little difference in the final results. For simplicity, we use the 8 corners of 3D bounding box as keypoints in our experiments.

\begin{table}
\caption{Ablation studies of different novelties on LINEMOD dataset. The metric we used to measure performance is ADD(-S) metric. ``SP" means 3D sphere, ``EVF" means embedding vector feature, and ``RRE" denotes rotation residual estimator. }
\vspace{-5pt}
\begin{center}
\begin{tabular}{|c|c|c|c|c|c|}
\hline
 Method &SP& EVF & RRE &  Acc  & Speed(fps)\\
\hline\hline
EXP1 &$\times$ &$\times$&$\times$ &93.4\%&25\\
EXP2 &$\checkmark$ &$\times$&$\times$ &95.8\%&25\\
EXP3 &$\checkmark$ &$\checkmark$&$\times$ &98.4\%&23\\
EXP4 &$\checkmark$ &$\checkmark$&$\checkmark$ &98.7\%&23\\
\hline
\end{tabular}
\end{center}
\label{tab:abla_evf}
\vspace{-10pt}
\end{table}

\begin{table}
\caption{Ablation studies of different keypoints parameters on LINEMOD dataset. The metric we used to measure performance is ADD(-S) metric. \textbf{BBX-8} means using the 8 corners of 3D bounding box as keypoints. \textbf{FPS-K} denotes that we adopt $K$ keypoints generated by the FPS algorithm.}
\vspace{-5pt}
\begin{center}
\begin{tabular}{|c|c|c|c|c|}
\hline
 Method &BBX-8& FPS-4 & FPS-8 & FPS-12\\
\hline\hline
 Acc & 98.7\% &98.5\%&98.4\% &98.6\%\\ 
\hline
Speed (fps) &23 &23&23 &23\\ 
\hline
\end{tabular}
\end{center}
\vspace{-15pt}
\label{tab:abla_kps}
\end{table}

\subsection{Generalization performance}
\label{sec:conver}
\begin{figure}
\begin{center}
\begin{tabular}{cc}
{{\includegraphics[width=3.8cm, height=3cm]{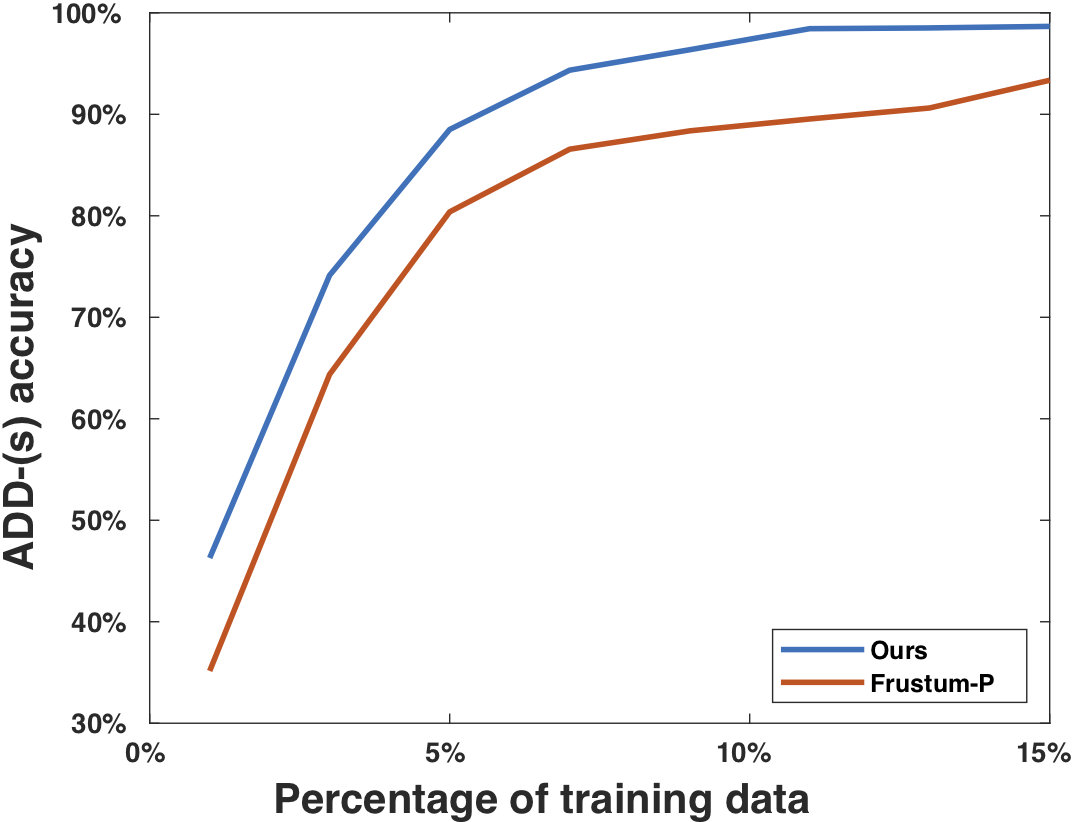}}}&
{{\includegraphics[width=3.8cm,height=3cm]{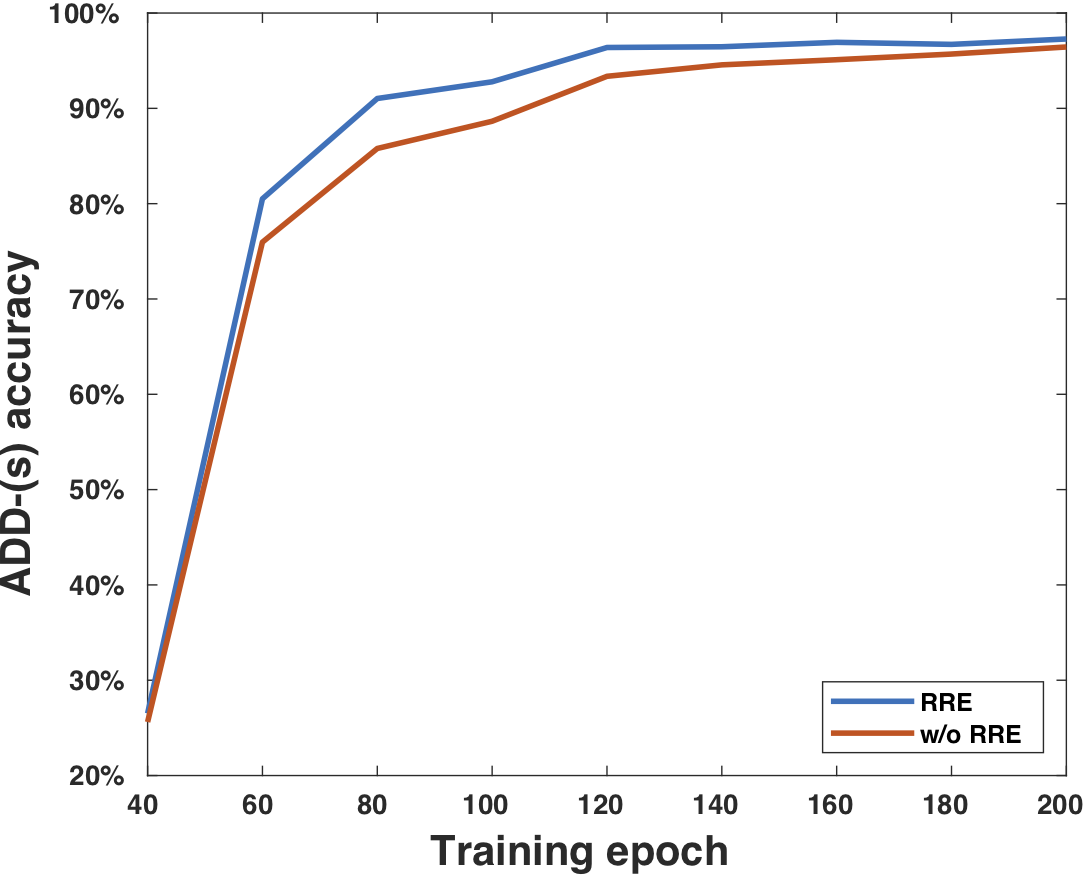}}}\\
(a)&(b)\\
\end{tabular}
\end{center}
\caption{\textbf{Visualization of method performance on LINEMOD dataset.} (a) {Influence of training data size using the ADD metric. When using the same training size, compared to Frustum-P \cite{Qi_2018_CVPR}, our method improves the performance significantly. For simplicity, here we provide ground truth 2D bounding box and randomly choose an object point as 3D sphere center for evaluation}. (b) As the rotation localization network converging, the impacts of rotation residual estimator (RRE) decreases. }
\label{fig:gen_per}
\end{figure}

In this section, we evaluate the generalization performance of our G2L-Net. We gradually reduce the size of training data to see how the performance of the algorithm can be affected on LINEMOD dataset. From Figure \ref{fig:gen_per} (a), we can see that even only $5\%$ of the training data, which is 1/3 of the normal setting, is used for the network training, the performance (88.5\%) is still comparable. 

\begin{table*}[t!]
\caption{6D pose estimation accuracy on LINEMOD dataset. We use ADD metric to evaluate the methods. For symmetric objects $Egg\ Box$ and $Glue$, we use ADD-S metric. Note that, we summarize the pose estimation results reported in the original papers on LINEMOD dataset.}
\begin{center}
\resizebox{165mm}{35mm}{
\begin{tabular}{|l|c|c|c|c|c|c|c|}
\hline
Method & \tabincell{c}{PVNet \cite{peng2018pvnet}}&\tabincell{c}{PoseCNN +\\ DeepIM \cite{Xiang2017, Li_2018_ECCV}}&DPOD \cite{zakharov2019dpod} & Frustum-P \cite{Qi_2018_CVPR} & Hinterstoisser \cite{hinterstoisser2016going}  & DenseFusion \cite{wang2019densefusion} & Ours\\
\hline\hline
Input& RGB & RGB&RGB & RGB+Depth&  Depth  & RGB+Depth &   RGB+Depth   \\
\hline
Refinement& $\times$ & $\checkmark$&$\checkmark (\times)$ &$\times$&  $\checkmark$  & $\checkmark(\times)$ &   $\times$  \\ 
\hline
Ape& 43.6\% & 77.0\% &87.7\% (53.3\%)& 85.5\%&  \textbf{98.5\%}  & 92.3\% (79.5\%)& {96.8}\%   \\ 

Bench Vise & \textbf{99.9\%} &97.5\%&98.5\% (95.3\%)&{93.2\%}&    99.0 \%& 93.2\%(84.2\%)& 96.1\% \\ 

Camera &86.9\%&93.5&96.0\% (90.4\%)&90.0\%&  \textbf{99.3\%}& 94.4\%(76.5\%)&  98.2\%\\ 

Can &95.5\%&96.5\%&99.7\% (94.1\%)&91.4\%& \textbf{98.7\%} & 93.1\%(86.6\%)&  {98.0\%}\\ 

 Cat &79.3\%&82.1\%&94.7\% (60.4\%)&96.5\%& \textbf{99.9\%} &  96.5\%(88.8\%)&  {99.2\%}\\ 
 
 Driller &96.4\%&95.0\%&98.8\% (97.7\%)&96.8\%&93.4\% & 87.0\%(77.7\%)&   \textbf{99.8\%}\\ 

 Duck &52.6\%&77.7\%&86.3\% (66.0\%)&82.9\%&\textbf{98.2\%} & 92.3\%(76.3\%)&  97.7\%\\ 

 Egg Box &99.2\%&97.1\%&99.9\% (99.7\%)&{99.9\%}&98.8\% &  {99.8\%}(99.9\%)&  \textbf{100\%}\\ 

 Glue &95.7\%&99.4\%&96.8\% (93.8\%)& {99.2\%}&75.4\% & \textbf{100\%} (99.4\%)&  \textbf{100\%}\\
 
 Hole Puncher &81.9\%&52.8\%&86.9\% (65.8\%)&92.2\%&{98.1\%}  & 92.1\%(79.0\%)&   \textbf{99.0\%}\\ 

 Iron &{98.9\%}&98.3\%&\textbf{100\%} (99.8\%)&93.7\%&98.3\%  & 97.0\% (92.1\%)& { 99.3\%}\\ 

 Lamp & {99.3\%}&97.5\%&96.8\% (88.1\%)&98.2\%&96.0\% & 95.3\%(92.3\%)&   \textbf{99.5\%}\\ 

 Phone &92.4\%&87.7\%&94.7\% (74.2\%)&94.2\%&{98.6\%} & 92.8\%(88.0\%)& \textbf{98.9\%}\\ 
 \hline
  Speed(FPS) &25&5 & 33(40) &12& 8 & 16(20) & 23 \\ 
\hline
\hline
Average &86.3\%&88.6\%&95.2\% (83.0\%) &93.4\%&96.3 \% &94.3 \%(86.2\%)&\textbf{98.7 \%}\\
\hline
\end{tabular}
}
\end{center}
\label{tab:linemod}
\end{table*}

\begin{figure*}[t!]
\begin{center}
\begin{tabular}{cccc}
{{\includegraphics[width=3.9cm]{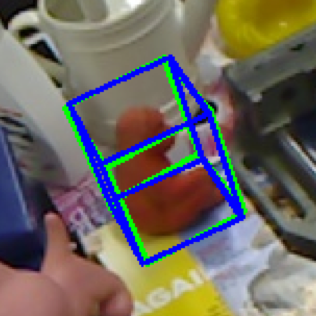}}}&
{{\includegraphics[width=3.9cm]{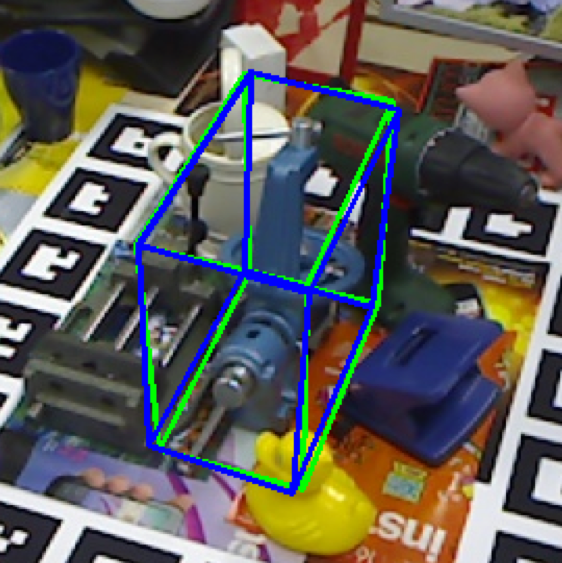}}}&
{{\includegraphics[width=3.9cm]{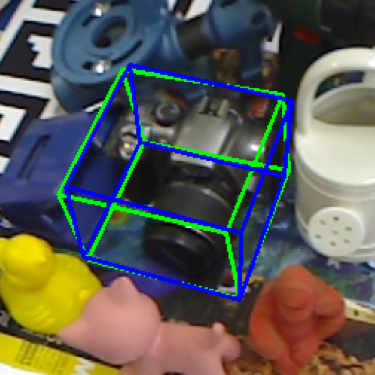}}}&
{{\includegraphics[width=3.9cm]{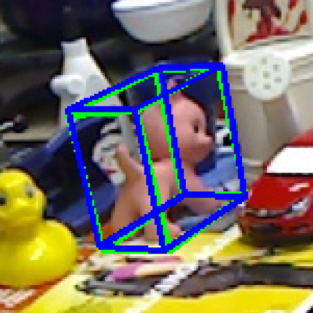}}}\\
{{\includegraphics[width=3.9cm]{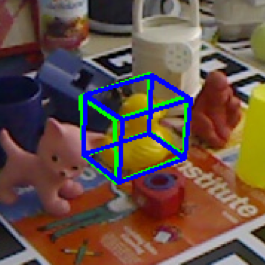}}}&
{{\includegraphics[width=3.9cm]{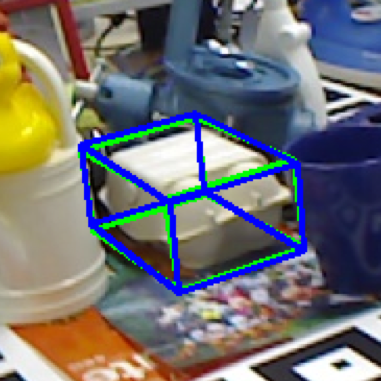}}}&
{{\includegraphics[width=3.9cm]{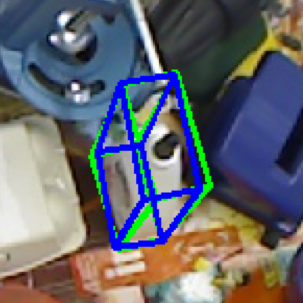}}}&
{{\includegraphics[width=3.9cm]{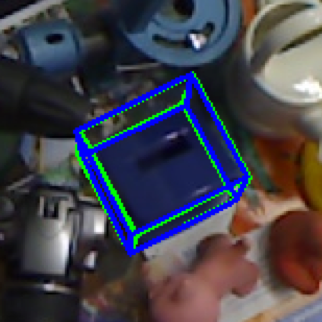}}}\\
\end{tabular}
\end{center}
\vspace{-5pt}
\caption{\textbf{Qualitative pose estimation results on LINEMOD dataset}. Green 3D bounding boxes denote ground truth. Blue 3D bounding boxes represent our results. Our results match ground truth well.}
\label{fig:linmod}
\end{figure*}

 \begin{table}[t!]
\caption{6D Pose estimation accuracy on the YCB-V dataset. We use ADD-S AUC metric to evaluate the methods.}
\vspace{-10pt}
\begin{center}
\resizebox{0.48\textwidth}{35mm}{
\begin{tabular}{|l|c|c|c|c|}
\hline
Method(RGB+Depth) &\tabincell{c}{PoseCNN \cite{Xiang2017} + ICP } & MCN \cite{Li_2018_ECCV} & \tabincell{c}{DenseFusion \cite{wang2019densefusion}\\ (no refinement)} & Ours\\
\hline\hline
002\_master\_chef\_can& 95.8\%&  \textbf{96.2}\%  & {95.2}\% &   {94.0}\%   \\ 

003\_cracker\_box &{91.8\%}&    90.9 \%& \textbf{92.5}\%& 88.7\% \\ 

004\_sugar\_box &\textbf{98.2}\%&  {95.3\%}& 95.1\%&  96.0\%\\ 

005\_tomato\_soup\_can  &94.5\%&\textbf{97.5}\% & 93.7\%&  {86.4}\%\\ 

 006\_mustard\_bottle &\textbf{98.4}\%& {97.0\%} &  95.9\%&  {95.9\%}\\ 
 
 007\_tuna\_fish\_can  &\textbf{97.1}\%&95.1\% & 94.9\%&   {96.0\%}\\ 

 008\_pudding\_box  &\textbf{97.9}\%&{94.5\%} & 94.7\%&  93.5\%\\ 

009\_gelatin\_box  &\textbf{98.8}\%&96.0\% &  {95.8\%}&  96.8\%\\ 

 010\_potted\_meat\_can  & {92.8\%}&\textbf{96.7}\% & {90.1\%} &  {86.2\%}\\ 

011\_banana &\textbf{96.9}\%&{94.4\%}  & 91.5\%&   {96.3}\%\\ 

 019\_pitcher\_base  &\textbf{97.8}\%&96.2\%  & 94.6\% & { 91.8\%}\\ 

 021\_bleach\_cleanser &\textbf{96.8}\%&95.4\% & 94.3\%&   {92.0\%}\\ 

 024\_bowl& 78.3\%&82.0\% & {86.6}\%&\textbf{ 86.7}\%\\ 
 
 025\_mug& 95.1\%&\textbf{96.8}\% & {95.5}\%& {95.4}\% \\
 035\_power\_drill& \textbf{98.0}\%&93.1\% & 92.4\%& {95.2\%} \\
 036\_wood\_block& 90.5\%&\textbf{93.6}\% & 85.5\%& {86.2\%}\\
 037\_scissors & 92.2\%&94.2\% & \textbf{96.4}\%& {83.8\%}\\
 040\_large\_marker& \textbf{97.2}\%&95.4\% & {94.7}\%& {96.8\%}\\
 051\_large\_clamp & 75.4\%&93.3\% & 71.6\%& \textbf{94.4}\%\\
 052\_extra\_large\_clamp& 65.3\%&90.9\% & 69.0\%& \textbf{92.3}\%\\
 061\_foam\_brick& \textbf{97.1}\%&95.9\% & 92.4\%& {94.7\%}\\
\hline
\hline
Average &93.0\%&\textbf{94.3} \% &91.2 \%& 92.4 \%\\
\hline
Speed (fps) &$<$ 0.1 & $<$ 10  & 20& \textbf{21} \\
\hline
\end{tabular}}
\end{center}
\vspace{-15pt}
\label{tab:ycbv}
\end{table}

 \begin{figure}[t!]
\begin{center}
\begin{tabular}{cc}
{{\includegraphics[width=3.8cm]{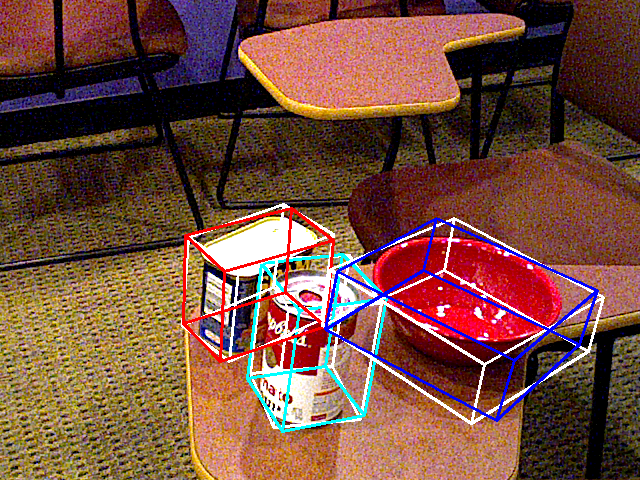}}}&
{{\includegraphics[width=3.8cm]{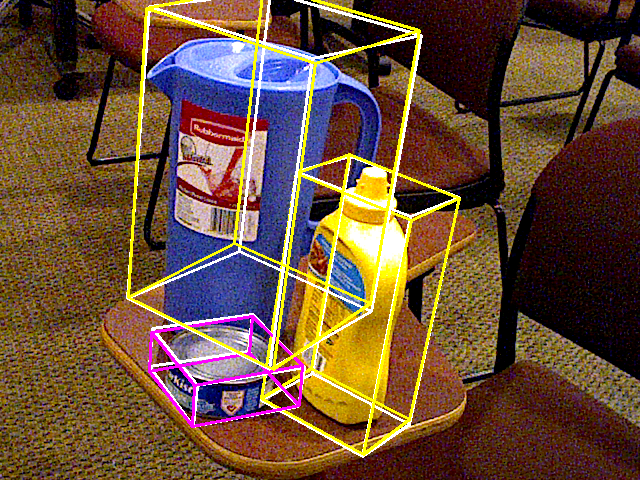}}}\\
\end{tabular}
\end{center}
\caption{\textbf{Visualizing pose estimation results on YCB-Video}. White 3D bounding boxes are ground truth. Colorful 3D bounding boxes represent our results. For different objects, our prediction matches ground truth well.}
\label{fig:ycb_v}
\end{figure}


\subsection{Comparison with the state-of-the-art methods}

\noindent\textbf{Object 6D pose estimation on LINEMOD: }
 Same as other state-of-the-art methods, we use 15\% of each object sequence to train and the rest of the sequence to test on LINEMOD dataset. In Table \ref{tab:linemod}, we compare our method with state-of-the-art RGB and RGB-D methods. The numbers in brackets are the results without post-refinement.
 We use Frustum-P \cite{Qi_2018_CVPR} as our baseline. We re-implement Frustum-P to regress 3D bounding box corners of the objects. From Table \ref{tab:linemod}, we can see that our method outperforms the baseline by 5.4\% in ADD accuracy and runs 2 times faster than the baseline method. Comparing to the second-best method \cite{hinterstoisser2016going} that using depth information, our method outperforms it by 2.4\% in ADD accuracy and runs about 3 times faster than it. Although DPOD and PVNet are faster than our method, they only take RGB image as input. When using depth information, our method achieves the fastest inference speed. In Figure \ref{fig:linmod}, we provide a visual comparison of predict pose versus ground truth pose. \\

\noindent\textbf{Object 6D pose estimation on YCB-Video: }

Different from LIMEMOD dataset, in YCB-Video dataset, each frame may contain multiple target objects. Our method can also estimate 6D pose for multiple objects in fast speed. 
Table \ref{tab:ycbv} compares our method with other state-of-the-art methods \cite{Xiang2017,Li_2018_ECCV,wang2019densefusion} on YCB-Video dataset under ADD-S AUC metric. From Table \ref{tab:ycbv}, we can see that our method achieves a comparable accuracy (92.4\%) and is the fastest one (21fps) among all comparisons. In Figure \ref{fig:ycb_v}, we also provide visualization results on this dataset.

\subsection{Running time}
For a single object, given a $480 \times 640 $ RGB-D image, our method runs at $23$fps on a PC environment (an Intel i7-4930K 3.4GHz CPU and one GTX 1080 Ti GPU).
Specifically, the 2D detector takes $11ms$ for object location, and pose estimation part which includes translation localization and rotation localization takes  $32ms$. The rotation residual estimator takes less than $1ms$.

\section{Conclusion}
In this paper, we propose a novel real-time 6D object pose estimation framework. Our G2L-Net decouples the object pose estimation into three sub-tasks: global localization, translation localization and rotation localization with embedding vector features. In the global localization, we use a 3D sphere to constrain the 3D search space into a more compact space than 3D frustum. Then we perform 3D segmentation and object translation estimation. We use the 3D segmentation mask and the estimated object translation to transfer the object points into local coordinate space. Since viewpoint information is more evident in this canonical space, our network can better capture the viewpoint information with our proposed point-wise embedding vector features. In addition, to fully utilize the viewpoint information, we add the rotation estimation estimator, which learns the residual between the estimated rotation and ground truth. In experiments, we demonstrate that our method achieves state-of-the-art performance in real-time. 

Although our G2L-Net achieves state-of-the-art performance, there are some limitations to our framework. First, our G2L-Net relies on a robust 2D detector to detect the region of interest.
Second, while our network exploits viewpoint information from the object point cloud, the texture information is not well adopted. In future work, we have a plan to overcome these limitations.
\newline

\noindent
\textbf{{Acknowledgement}}
We acknowledge MoD/Dstl and EPSRC (EP/N019415/1) for providing the grant to support the UK academics involvement in MURI project.

\newpage
{\small
\bibliographystyle{ieee_fullname}
\bibliography{egbib}
}

\end{document}